\documentclass[letterpaper, 10 pt, conference]{ieeeconf}  

\usepackage{epsfig}
\usepackage{graphicx}

\IEEEoverridecommandlockouts                              

\title{\LARGE \bf
DADA-2000: Can Driving Accident be Predicted by Driver Attention? Analyzed by A Benchmark
}

\author{Jianwu Fang$^{1,2}$, Dingxin Yan$^{1}$, Jiahuan Qiao$^{1}$, Jianru Xue$^{2}$, He Wang$^{1}$ and Sen Li$^{1}$
\thanks{*This work was supported by the Natural Science Foundation of China (No. 61751308, 61603057, and 61773311), and in part supported by the China Postdoctoral Science Foundation (No. 2017M613152)}
\thanks{$^{1}$J. Fang, D. Yan, J. Qiao, H. Wang and S. Li are with the School of Electronic and Control Engineering, Chang'an University, Xi'an, China; J. Fang is also with the Institute of Artificial Intelligence and Robotics, Xi¡¯an Jiaotong University, Xi'an, China
        {\tt\small fangjianwu@chd.edu.cn}.}%
\thanks{$^{2}$J. Xue is with the Institute of Artificial Intelligence and Robotics, Xi¡¯an Jiaotong University, Xi'an, China
        {\tt\small jrxue@mail.xjtu.edu.cn}.}%
}

\begin{document}

\maketitle
\thispagestyle{empty}
\pagestyle{empty}

\begin{abstract}

Driver attention prediction is currently becoming the focus in safe driving research community, such as the DR(eye)VE project and newly emerged Berkeley DeepDrive Attention (BDD-A) database in critical situations. In safe driving, an essential task is to predict the incoming accidents as early as possible. BDD-A was aware of this problem and collected the driver attention in laboratory because of the rarity of such scenes. Nevertheless, BDD-A focuses the critical situations which do not encounter actual accidents, and just faces the driver attention prediction task, without a close step for accident prediction. In contrast to this, we explore the view of drivers' eyes for capturing multiple kinds of accidents, and construct a more diverse and larger video benchmark than ever before with the driver attention and the driving accident annotation simultaneously (named as \texttt{DADA-2000}), which has $2000$ video clips owning about $658,476$ frames on $54$ kinds of accidents. These clips are crowd-sourced and captured in various occasions (highway, urban, rural, and tunnel), weather (sunny, rainy and snowy) and light conditions (daytime and nighttime). For the driver attention representation, we collect the maps of fixations, saccade scan path and focusing time. The accidents are annotated by their categories, the accident window in clips and spatial locations of the crash-objects. Based on the analysis, we obtain a quantitative and positive answer for the question in this paper.
\end{abstract}

\section{Introduction}

On the journey to truly safe driving, human-centric assistive driving systems are still the focus by exploring the agile, dexterous driving experience of human beings. Within the human knowledge exploitation, driver attention \cite{palazzi2018predicting,rasouli2018joint} has the undeniable role to optimize the visual scene search (e.g., foveal vision \cite{Perry2002Gaze}) with a saving of computational resources when driving in complex real-world \cite{guo2018safe}. In the meanwhile, driver attention is an important way to interact with the surroundings, and joint attention exploration \cite{rasouli2018joint,rasouli2017agreeing} between drivers and other road users can promote the intention prediction for safe-driving. In particular, because of the fast reaction of humans, the eye movement can respond immediately for critical driving situations. Therefore, in order to promote the development of truly safe-driving, we will study the relationship between driver attention prediction and actual accident prediction.
\begin{figure}[t]
 \setlength{\abovecaptionskip}{0.0cm}
\centering
\includegraphics[width=\linewidth]{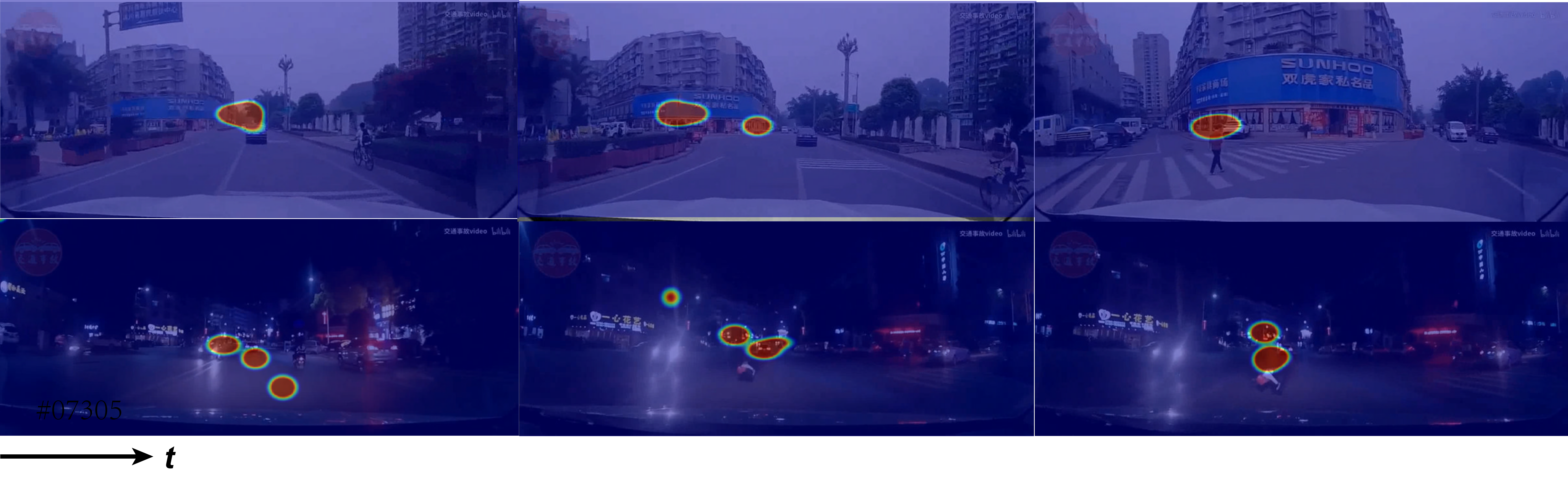}
    \caption{\small{The attention process of five observers from a normal situation to critical situation.}}
\label{fig1}
\end{figure}

In complex driving scenes, the simulation of driver attention is rather challenging and highly subjective because of the various attributes of persons, such as driving habit, driving experience, age, gender, personal preference, culture, and so on. Therefore, the fixations of different drivers vary considerably on the same scene, as demonstrated in Fig.~\ref{fig1} with two typical pedestrian crossing situations. Over decades, simulation of driver attention has been emphasised on driving behavior understanding and is now formulated by computer vision techniques in the autonomous driving or intelligent assistive driving systems. The goal of this task is to seek the answers for the question ``\emph{Where and what should we look when driving in different environments?}", and identify the regions of interest for further processing \cite{DBLP:journals/tits/DengYLY16,DBLP:conf/ivs/PalazziSCAC17}.

Recently, some attempts are launched, and the most typical project is DR(eye)VE \cite{palazzi2018predicting}, which collected $555,000$ video frames with driver attention maps in a practical driving car (we call it \emph{in-car} collection). However, the eye fixation data in DR(eye)VE project focused on the sunny and unobstructed road scenes, where very few critical situations appeared, and the attention collection stood on one driver's view. This setting cannot reflect the common sense in driving. Hence, Berkeley DeepDrive Laboratory recently launched a project to collect the driver attention for critical situations, where the dataset is called Berkeley DeepDrive Attention (BDD-A) \cite{xia2017predicting} with $1232$ videos. Because of the rarity of the critical situations in practical driving, BDD-A designed an in-lab collection protocol. BDD-A claimed that in-lab collection is actually better than in-car collection because observers in lab are more focused without 1) the disturbance of surroundings in open environment and 2) extra manoeuvres for car controlling inevitably introducing the attention distraction.

Although BDD-A realizes the importance of driver attention in critical situations, there are three major aspects that we think can be boosted for truly safe-driving. (1) For a convenience to build learning models, BDD-A treats the attention of multiple observers on buildings, trees, flowerbeds, and other unimportant objects as noise and wash it out with eye movement average procedure. However, because of the highly subjectivity, these attention data maybe contextual for critical situation prediction. Seeing Fig. \ref{fig1} as an example, the fixations of observers tend to be close together when seeing the crash-object. We call it as a ``\emph{common focusing effect}". (2) The critical situations in BDD-A do not cause true accidents, where the attention simulation has not explored the dynamic process from critical situation to actual accidents (we call it as accident attention flow), which is needed for avoiding true accident. (3) BDD-A do not classify the braking events into different categories useful for fine-grained analysis.

In view of these aspects, this work contributes a new, larger and more diverse video benchmark with driver attention and driving accident annotation simultaneously. (we call it as \texttt{DADA-2000}). Because the accidents are rarer than the critical situations, we first collect crowd-sourced videos with $3$ millon frames. Then, we wash and divide them into $54$ kinds of accidents with respect to the their participants, where $658,476$ frames are selected out for further annotation. Second, the attention collection, similar to BDD-A, is conducted in lab with a carefully designed protocol. The driver attention data is collected by a SMI RED250 eyetracker, which contains the information of fixations (fixation maps, fixation duration, start time, end time, positions, start positions, end positions, dispersions, and pupil size). Third, we carefully annotated the locations of crash-objects and their temporal occurrence intervals. Totally, DADA-2000 contains $2000$ video clips, in which the scenes are fairly complex and diverse in weather conditions (sunny, snowy, and rainy), light conditions (daytime and nighttime), and occasions (highway, urban, rural, and tunnel).

With DADA-2000, we not only want to know ``\emph{Where and what should we look when driving in different environments?}", but also we want to seek the answers for a new question ``\emph{Can driving accident be predicted by driver attention?}". In addition, because we collected the attention flow from the normal situations to accidents, the more challenging question ``\emph{What causes the occurrence of an accident that we can see}?" can be explored in the future.

In a summary, this work has two \textbf{contributions}:

(1) We construct a new, larger and more diverse benchmark for driving accident prediction or detection, driver attention prediction problems than ever before, where each frame owns its corresponding attention data.

(2) We give an quantitative and positive answer for the question: Can driving accident by predicted by driver attention? This work not only provides a new solution for driving accident prediction or detection, but also pushes the driver attention one step further for safe-driving.

\section{Related Works}
Since this work concentrates on the driver attention mechanism understanding when encountering accidents, two main domains are involved,  \emph{driver attention prediction} and \emph{driving accident prediction} with computer vision techniques.

\textbf{Driver Attention Prediction.} Driving has clear destination and path. Therefore, for the driver attention studying, it manifestly belongs to the task-specific attention field. Over decades, safety of self-driving cars has been strengthened by the robust perception of human-designated information, such as traffic signs, pedestrians, vehicles, road, as well as other kinds of traffic participants. Benefit from the progress of saliency computation models, driver attention that directly links the driving task and eye fixations is focused, but concentrates on the designated objects \cite{Xie2009Unifying,DBLP:journals/tits/WangHXYL17} for a long time. In order to mimic the real driver attention mechanism, the DR(eye)VE project \cite{palazzi2018predicting} was launched, and on this basis, several models based on deep neural networks \cite{DBLP:journals/tits/DengYLY16,DBLP:conf/ivs/TawariK17,DBLP:conf/ivs/PalazziSCAC17} were built for driver attention prediction. However, as aforementioned, there are some issues in this dataset. The major one in relevant to our work is that the scene is unobstructed and does not consider the accident situations. Beside DR(eye)VE, there were also some attempts \cite{DBLP:journals/tits/DengYLY16}, whereas the datasets in these attempts were annotated coarsely and cannot reflect the practically dynamic driving behavior. More recently, Berkeley DeepDrive Laboratory constructed a large-scale driver attention dataset in-lab focusing on the critical situations, named as BDD-A, and built a simple convolutional neural networks (CNN) to predict the driver fixations. BDD-A is the most relative one to our work, whereas it does not consider the dynamic attention process from the critical situations to actual accidents. In the meantime, they did not categorize the braking events into sub-classes, which may be more useful for avoiding certain accident. In this paper, we provide a new and larger driver attention benchmark considering a more diverse driving situations.
\begin{figure*}[t]
 \setlength{\abovecaptionskip}{0.0cm}
\centering
\includegraphics[width=0.95\linewidth]{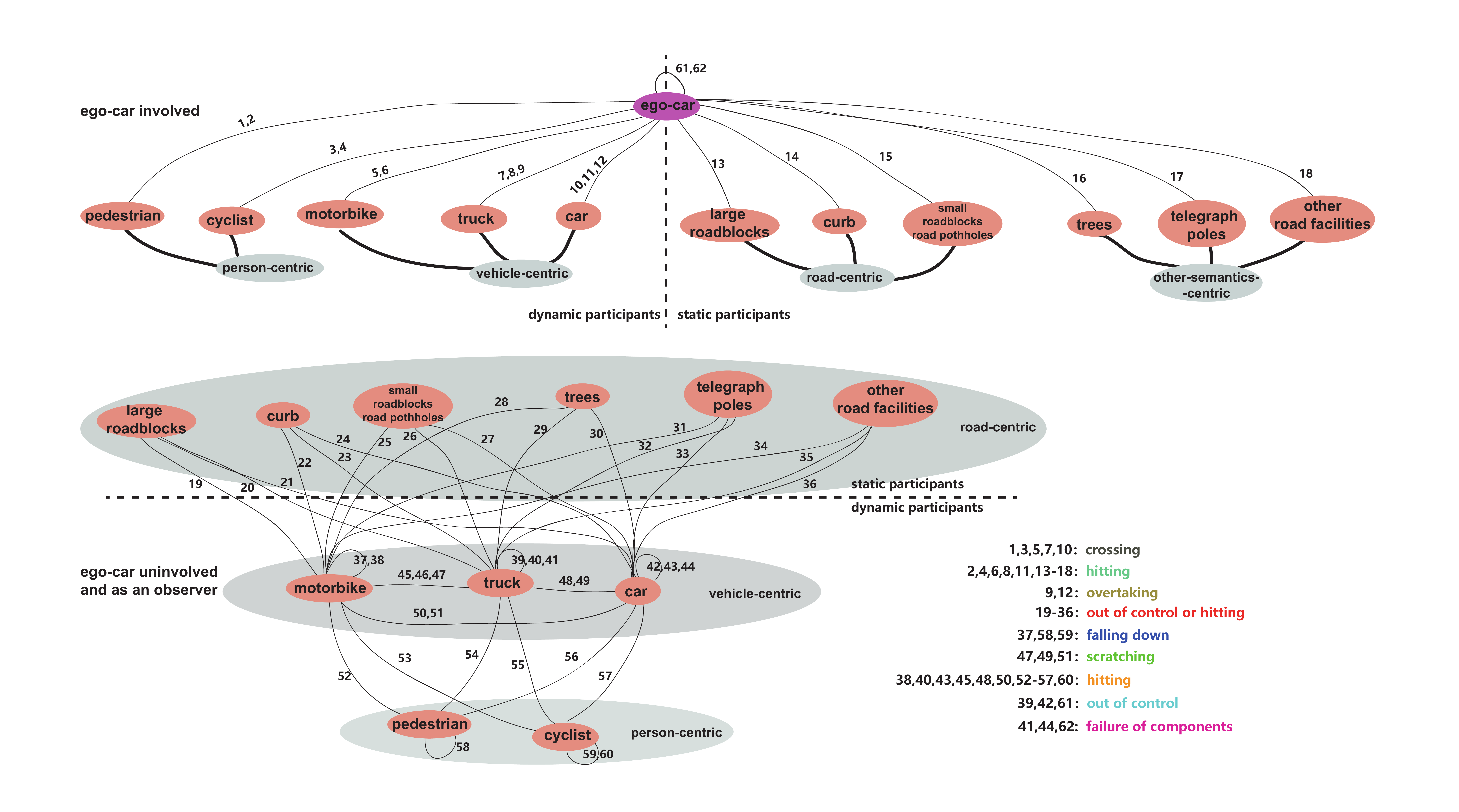}
    \caption{\small{The ego-car involved and ego-car uninvolved accident category graph in driving, where each kind of accident category is explained.}}
\label{fig2}
\end{figure*}

\textbf{Driving Accident Prediction.} Since this work involves the accident situations, and wants to devote ourselves to give an analysis between driving accident prediction and driver attention, we also review the literatures on accident prediction. Accident is generally a special anomaly in driving scenes. Compared with extensively studied anomaly in surveillance systems \cite{sultani2018real,yuan2015online}, accident has a more specific and clearer definition. Recent researches in computer vision have began to address the accident prediction or detection from different views. For instance, Kataoka \emph{et al.} \cite{suzuki2018anticipating,DBLP:conf/icra/KataokaSOMS18} and \emph{Chan et al.} \cite{chan2016anticipating} anticipated the traffic accident through adaptive loss and dynamic-spatial attention (DSA) recurrent neural network (RNN), respectively. Particularly, reference \cite{suzuki2018anticipating} contributed a large dashcam video dataset owning $6000$ videos temporally labeled, but has not been released. Yuan \emph{et al.} ~\cite{Yuan2017Incrementally} addressed the driving anomaly by incremental motion consistency measurement. Yao \emph{et al.} \cite{yao2019unsupervised} newly proposed a first-person video benchmark for driving accident detection, which contains $1500$ video clips with $208,166$ frames. Generally, the accident prediction in previous studies concentrate on the temporal interval prediction/detection, without a spatial crash object localization. In the meantime, the accurate prediction/detection of participants needs the expensively computed optical flow. More importantly, these methods cannot immediately respond the sudden accident because there is no enough time window for computation.

\section{DADA-2000}
\begin{figure}[t]
 \setlength{\abovecaptionskip}{0.0cm}
\centering
\includegraphics[width=\linewidth]{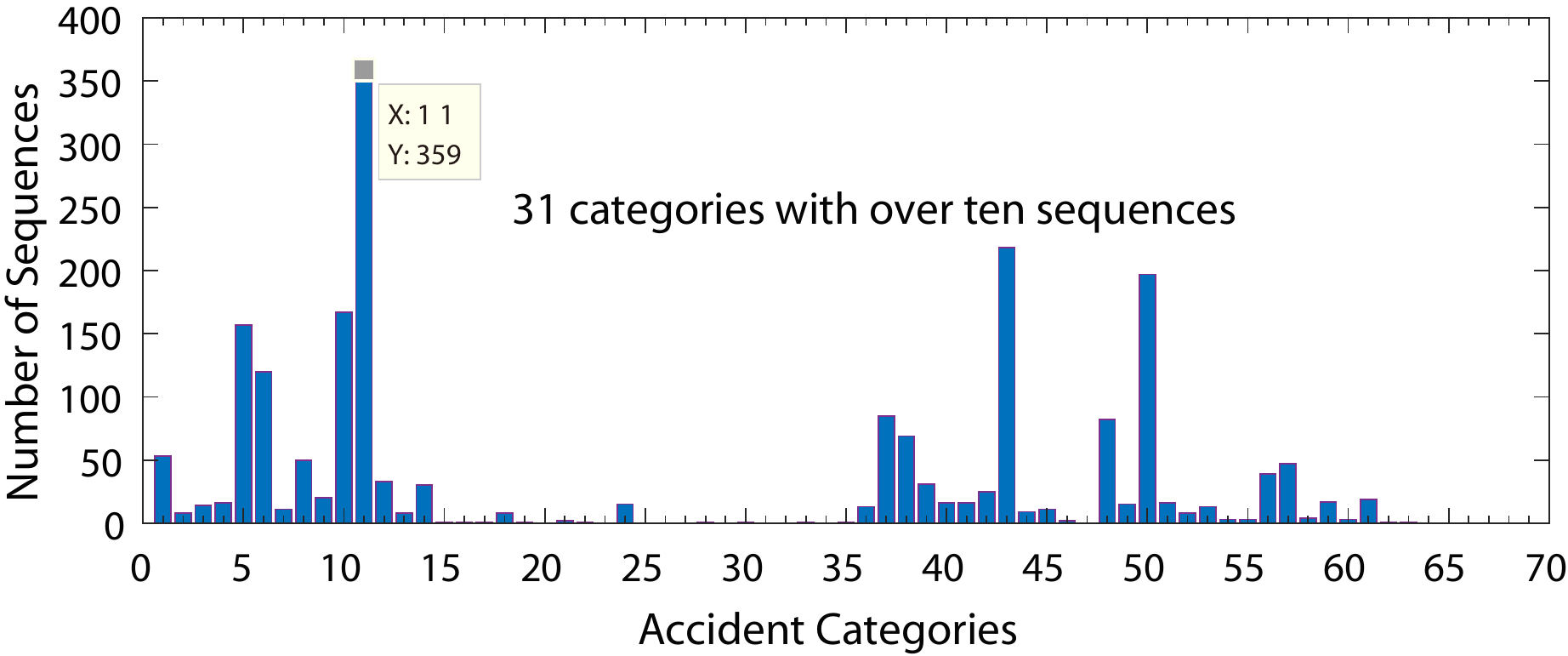}
    \caption{\small{The sequence amount distribution w.r.t. accident categories in DADA-2020.}}
\label{fig3}
\end{figure}

\begin{figure}
  \centering
  \includegraphics[width=\linewidth]{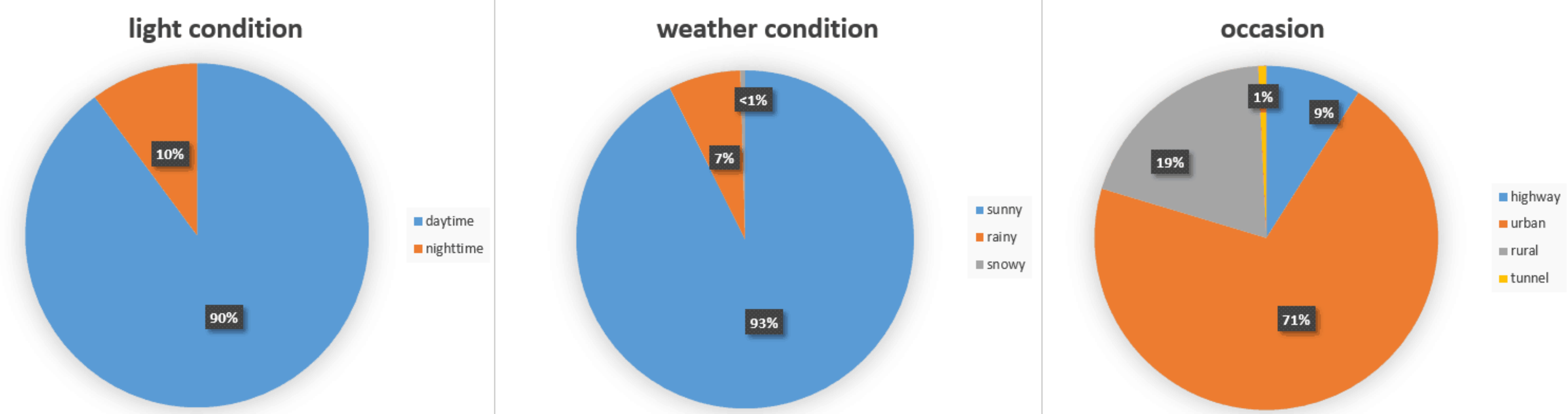}\\
  \caption{The distributions of environment attributes of DADA-2000. From left to right, the attributes are light conditions, weather conditions and scene occasions, respectively. Note that the weather condition statistics are based on the daytime because of a clear distinction.}
  \label{fig4}
  \vspace{-0.6cm}
\end{figure}
\begin{figure}[htpb]
 \setlength{\abovecaptionskip}{0.0cm}
\centering
\includegraphics[width=\linewidth]{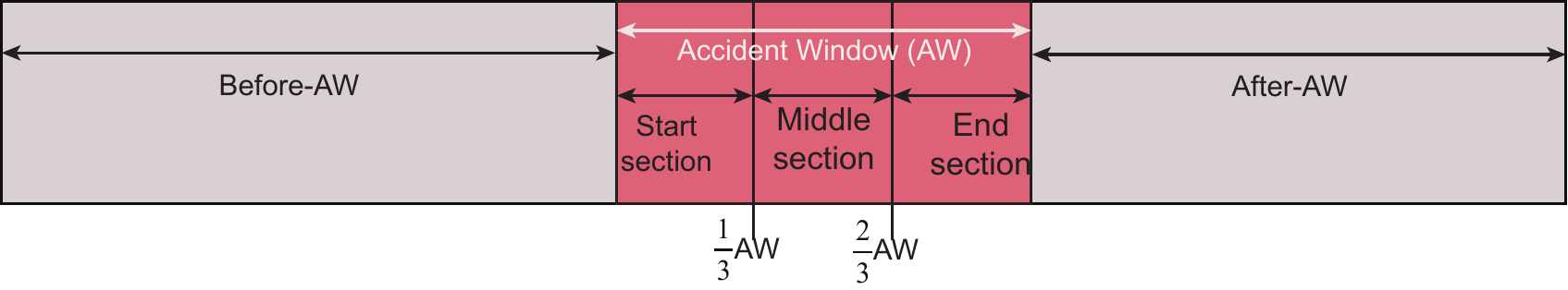}
    \caption{\small{Temporal partition of an accident video in this work.}}
\label{fig5}
\vspace{-0.6cm}
\end{figure}

In order to collect the accident videos as many as possible, we searched almost all the mainstream video websites, such as Youtube, Youku, Bilibili, iQiyi, Tencent, etc., and downloaded about \texttt{3 million} frames of videos. However, these videos have many useless typing masks. Therefore, we have conducted a laborious work for washing them, and obtain $658,476$ available frames contained in $2000$ videos with the resolution of $1456\times660$ (=6.1 hours with $30$ fps, over than DR(eye)VE). Different from the existing works with a strict trimming of frames \cite{chan2016anticipating,xia2017predicting} (such as the last ten frames as the accident interval) for accident annotation, we advocate a free presentation without any trimming work. In this way, the attention collection maybe more natural. We further divide these videos into $54$ kinds of categories based on the participants of accidents (pedestrian, vehicle, cyclist, motorbike, truck, bus, and other static obstacles, etc.). Among them, these $54$ categories can be classified into two large sets, ego-car involved and ego-car uninvolved. Specifically, the accident categories and their amount distribution of our benchmark is illustrated in Fig.~\ref{fig2} and Fig. ~\ref{fig3}, respectively. Because the accidents are rather diverse, we consider the accident situations as complete as possible in driving scene. Therefore, we provide $62$ categories denoted in Fig. \ref{fig2}, which is more than the collected $54$ classes. From this distribution, the ``\emph{ego-car hitting car}" takes the largest proportion. In addition, we also present the scene diversity of DADA-2000 by Fig. \ref{fig4}. From Fig. \ref{fig4}, we can see that because of the more frequent and diverse transit trip in daytime and urban scene than nighttime and other occasions, they show the highest occurrence rate of accidents.

In DADA-2000, we annotated the spatial crash-objects, temporal window of the occurrence of accidents, and collected the attention map for each video frame. Therefore, we will elaborate the constructing process of DADA-2000 in detail. Specifically, Fig. \ref{fig5} provides an illustration for temporal analysis of accident videos. For a video clip, we partition it into three main clips: the frame interval before the accident window (before-AW), accident window (AW) and the frame interval after the AW (after-AW). What is more, we further divide the AW into three sub-sections: start-section with the length of $\frac{1}{3}AW$, middle-section with the range from $\frac{1}{3}AW$ to $\frac{2}{3}AW$ and the end-section with remaining $\frac{1}{3}AW$.
Actually, in this work, AW not only contains the accident frames, but the temporal window prior to the accident is also contained. Therefore, this setting aims to explore whether driver attention can capture the crash-object or not once it appears in the start-section, i.e, driving accident prediction by driver attention. In addition, by observation, most of the actual accidents occur in the end section. Therefore, the start-section and middle-section are the golden interval for predicting accident. For AW determining, if half part of the object that will occur accident (we define it as crash-object in this paper) appears in the view plane, we set the frame as the starting point, and set the frame as the ending point if the scene returns a normal moving condition.
\begin{figure*}[t]
 \setlength{\abovecaptionskip}{0.0cm}
\centering
\includegraphics[width=0.8\linewidth]{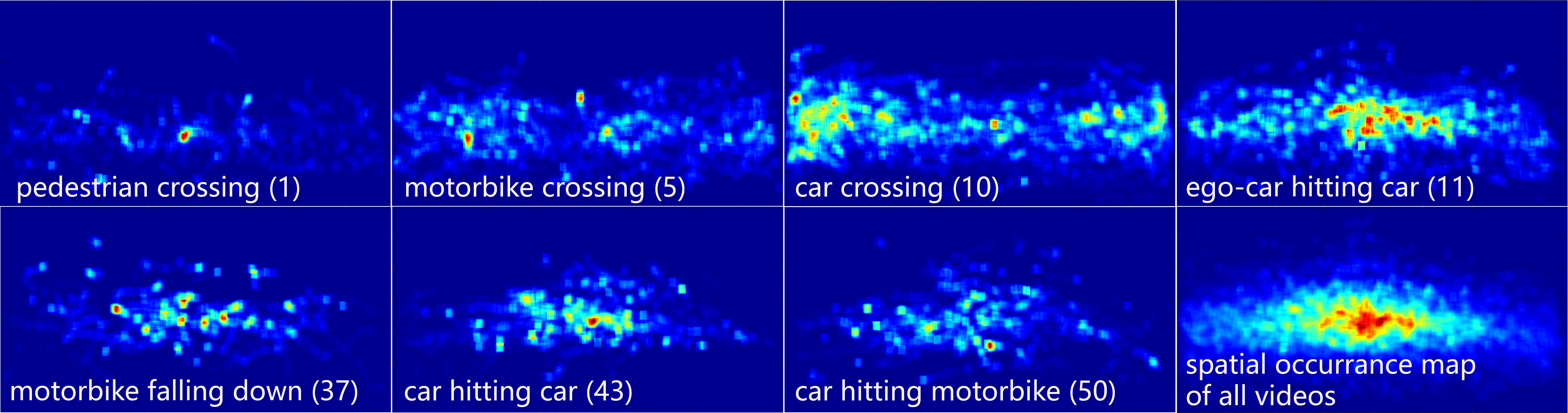}
    \caption{\small{The spatial occurrence maps of typical accident categories, as well as the occurrence map of DADA-2020, where the number in the bracket at the end of the accident class is the category index denoted in Fig. \ref{fig2}.}}
\label{fig6}
\end{figure*}

\begin{table}[htpb]
\centering
\caption{The temporal frame statistics of the number of frames and average frames of all videos, before-AW, AW, and after-AW, where AW represents the accident window.}
\begin{tabular}{|c|c|c|c|c|}

  \hline
  Statistics & total & before-AW & AW & after-AW \\
  \hline
  total frames & 650,000 & 315,154 & 131,679 & 211643 \\
 average frames & 325 & 157 & 66 & 106 \\
 percent (\%) & 100 &48.3 & 20.3 & 32.6\\
  \hline
\end{tabular}
\label{tab1}
\end{table}
\begin{table*}[htpb]
\centering
\caption{Attributes comparison of different driving accident datasets.}
\begin{tabular}{|c|c|c|c|c|c|}
  \hline
  Dataset & videos & accidents  & all frames  & typical participants & annotation type\\
  \hline
  ShanghaiTech \cite{DBLP:conf/iccv/LuoLG17} & 437 & 130  & 317,398 & bike, pedestrian & temporal\\
  Street Accidents (SA) \cite{chan2016anticipating} & 994 & 165 & 99,400 & car, truck, bike & temporal\\
  A3D** \cite{yao2019unsupervised}& 1500 & 1500 & 208,166 & car, truck, bike, pedestrian, animal & temporal\\
 \textbf{ DADA-2000}& 2000 & 2000 & 658,476 & car, truck, bike, pedestrian, animal, motorbike, static obstacles & spatial and temporal\\

  \hline
\end{tabular}
\label{tab2}
\end{table*}
\subsection{Temporal and Spatial Statistics of DADA-2000}

\subsubsection{Temporal Statistics}
In DADA-2000, the temporal frame statistics of before-AW, AW, and after-AW are presented in Table. \ref{tab1}. From these statistics, we find that the before-AW contains about $5$ seconds of time (30$fps$) in average and after-AW takes about 3.5 seconds of time. AW takes a percent of $20.3\%$ in each video averagely. Therefore, the frames of abnormal driving are rather fewer than the ones in normal driving.

\begin{table*}[htpb]
\centering
\caption{The attribute comparison of different driver attention datasets}
\begin{tabular}{|c|c|c|c|c|c|c|}

  \hline
  dataset &  rides & durations(hours) & drivers & gaze providers  & event & gaze patterns for each frame\\
  \hline
  DR(eye)VE \cite{palazzi2018predicting}& 74 & 6 & 8 & 8 & 464 braking events& attention map of single person\\
  BDD-A \cite{xia2017predicting} & 1232 & 3.5 & 1232  & 45 & 1427 braking events& average attention map of multiple observers\\
  \textbf{DADA-2000} & 2000 & 6.1 & 2000 & 20 & 2000 accidents (54 categories)&raw attention maps of multiple observers\\

  \hline
\end{tabular}
\label{tab3}
\end{table*}
\subsubsection{Spatial Statistics}
 Beside the temporal statistics, we also visualize the occurrence map of crash-object locations for some typical categories denoted in Fig. \ref{fig2}. These maps are obtained by summarizing the crash-object locations in each frame of certain accident category. For a clear demonstration, we expand the location point of crash-object with a rectangle neighborhood having the size of $14\times14$, and set the value within it as 1, and 0 for vice versa. Then, we normalize the summarized occurrence maps to the range of [0,1] with the min-max normalizer. From these maps, we can observe that the crash-object locations tend to the middle of the field of vision (FOV) for most of accident categories. However, \emph{pedestrian crossing}, \emph{motorbike crossing} and \emph{car crossing} are dispersive, and converge to the sides of FOV. Particularly, the cars with the larger velocity than motorbikes and pedestrians has the larger convergence degree to the FOV sides. On the contrary, locations of \emph{hitting} behaviour converge to the middle of FOV, and show a stronger degree of convergence with the increasing of the target velocity. Therefore, although the crash-object locations demonstrate a convergence of FOV, different accident categories with differing participants have diverse occurrence patterns of locations. This can be used for model designing with spatial context cueing for accident prediction or detection in future.

We also compare our DADA-2000 with the state-of-the-art datasets for driving accident detection or prediction in Table.~\ref{tab2}. From Table.~\ref{tab2}, we can observe that our DADA-2000 has more diverse scenarios and is more complex for driving accident analysis. This will provide a new and larger platform for driving accident detection or prediction problem.

\subsection{Attention Collection}
\subsubsection{Protocols}
Because of the rarity of the accident in practical driving, in our attention collection protocol, we employed $20$ volunteers with at least $3$ years of driving experience. The eye-tracking movement data were recorded in a lab by a SMI RED250 desktop-mounted infrared eye tracker with $250$ Hz, in conjunction with an iView X script. In order to approach the real driving scene, we weaken the lighting of the lab to reduce the impact of surroundings, by which only the computer screen is focused. In addition, we asked the volunteers to be relaxed and imagine that they are driving real cars. For avoiding the fatigue, we let each volunteer watch $40$ clips each time which are combined as a single long sequence with about $7$ minutes. Each clip was viewed at least by $5$ observers. It is worthy noting that, we ensure that the $40$ clips belong to the same category as much as possible, so as to prevent chaotic attention.

For collecting the attention data, we launched the \emph{no-priori} and \emph{with-priori} experiments. We ask each volunteer watch the same video twice, where in the first time the volunteers only know that they need to watch a video in driving scene (no priori), whereas they are told that they need to find the crash-object (with priori). By this experiment, we can analyze the unconscious crash attention and conscious crash attention, respectively. For our DADA-2000, we guarantee that each video is watched by at least five observers. For the attention map of a frame, there is a parameter determining the temporal window which aggregates the attentions within it to generate an attention map for a frame ($1$ second was utilized in DR(eye)VE). This setting can reserve the dynamic attention process in a small temporal window, but not be conducive to the crash-object localization.
\subsubsection{Statistics}
In this work, we save the attention maps as two types.
1) \emph{Type1}: We recorded the common attention maps, where an attention map contains five observers' fixations without temporal aggregation. Note that, different from the works \cite{xia2017predicting,palazzi2018predicting}, we do not average the attention fixations of observers and maintain them in the same frame because of their subjectivity. We can observe that the fixations get close to the appeared crash-object, and vice versa for normal driving scenes. We call this phenomena as ``\emph{common focusing effect}". 2) \emph{Type2}: We also recorded the attention maps of all the observers individually, where one-second of temporal window is used to obtain an aggregated attention map like DR(eye)VE project, in which the dynamic attention process can be reserved and useful for dynamic attention flow analyzing. These two types are all recorded in $30fps$. We capture the attention data for all of the frames in our DADA-2000. The attribute comparison with other state-of-the-art driver attention datasets is presented in Table.~\ref{tab3}. From this comparison, our DADA-2000 is more diverse, and contributes a new benchmark for driver attention prediction. Note that, our DADA-2000 will be released in future.
\begin{figure}[t]
 \setlength{\abovecaptionskip}{0.0cm}
\centering
\includegraphics[width=0.9\linewidth]{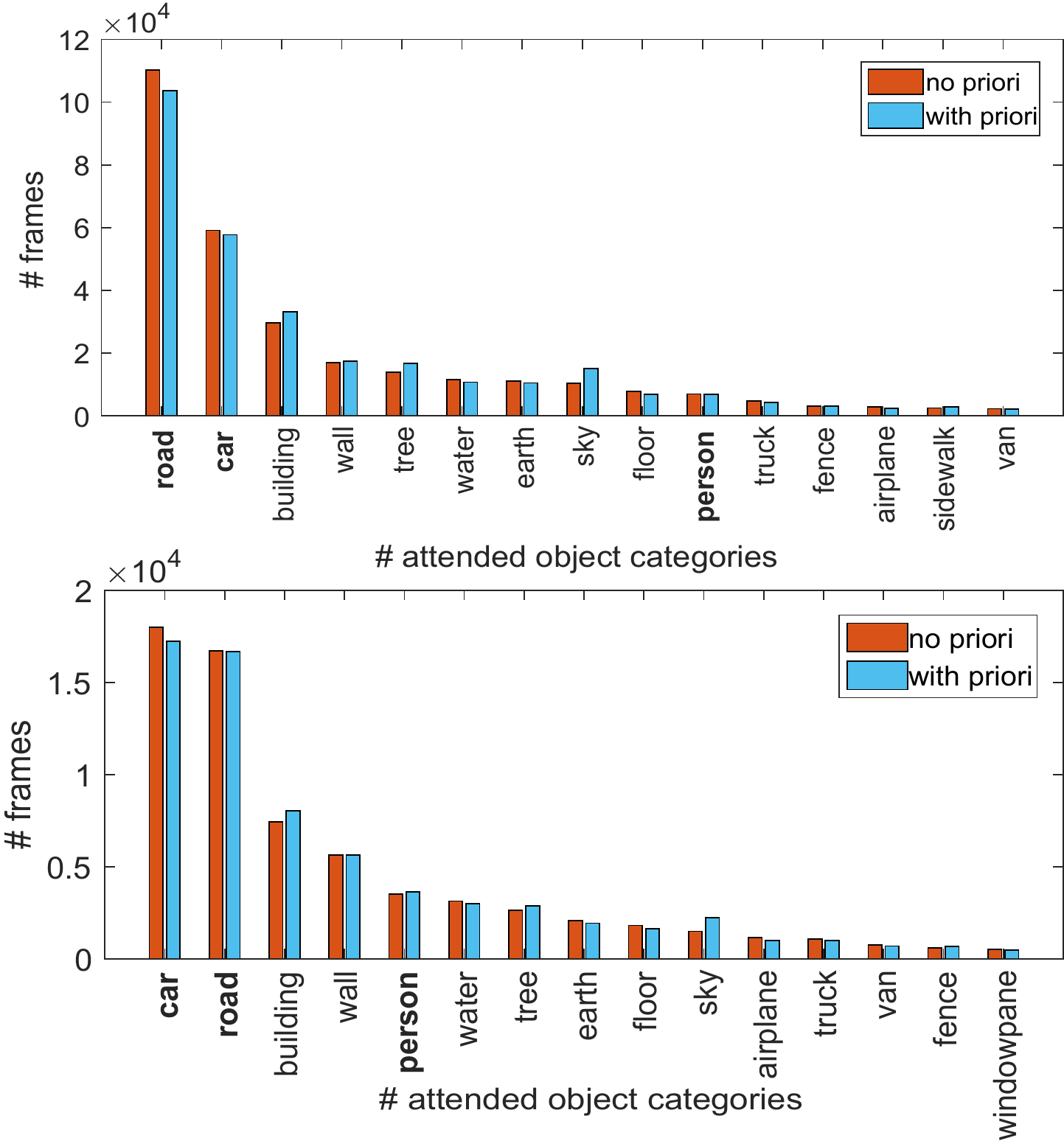}
    \caption{\small{Analysis of attended object categories: (Top row) normal driving situations (before- and after-AW) and (Bottom row) abnormal driving situations (AW).}}
\label{fig7}
\end{figure}

\section{Can Driving Accident be Predicted by Driver Attention?}
In this work, the most important purpose is to explore the question: Can driving accident be predicted by driver attention? The behind motivation is that, for the driving accident prediction or detection, current research paradigms often design and train complex models. However, the performance of previous works are still largely insufficient for safe-driving \cite{yao2019unsupervised}. Hence, we want to seek new solution for this task. Inspired by previous attempts \cite{rasouli2018joint}, driver attention is a powerful mechanism for object detection, especially for the sudden motion change. However, there is no promising studies with large-scale attention benchmark for driving accident analysis, where sudden motion changes are frequent.

\subsection{Attended Object Categories}
The crash objects in driving scene are usually the dynamic participants: such as the cars, pedestrians, motorbikes and other moveable objects. Therefore, these object categories should be attended more in abnormal driving situations. In addition, we also want to find the role of the priori for the crash-object localization when observing. As aforementioned, we conducted no-priori and with-priori experiments on all video clips. Hence, in Fig.~\ref{fig7}, we exhibit their analysis in parallel. Note that, the semantic annotation for our DADA-2000 is rather laborious, we get the semantic map of each frame by Deeplab-V3 model \cite{DBLP:journals/corr/ChenPSA17} pre-trained on ADE20 semantic segmentation dataset \cite{zhou2017scene}. In ADE20K, they defined $150$ object categories. In Fig.~\ref{fig7}, we demonstrate the object categories with top fifteen proportions.

From the analysis, we can see that the road and cars are attended more positive than other object categories. However, person shows a manifest shift from the tenth one to the fifth, and cars shift to the first column in abnormal driving situations. In other words, drivers attend the vulnerable road users more largely in abnormal driving situations than the ones in normal driving situations. In addition, we find that the priori for attention collection has little influence on the attending of object. Actually, human eyes are sensitive for the sudden motion change whatever the priori exists or not. These shifts can give a positive answer for the helpfulness of driver attention when predicting driving accident to some extent. However, this conclusion still needs more evidence.
\begin{figure}[htpb]
 \setlength{\abovecaptionskip}{0.0cm}
\centering
\includegraphics[width=\linewidth]{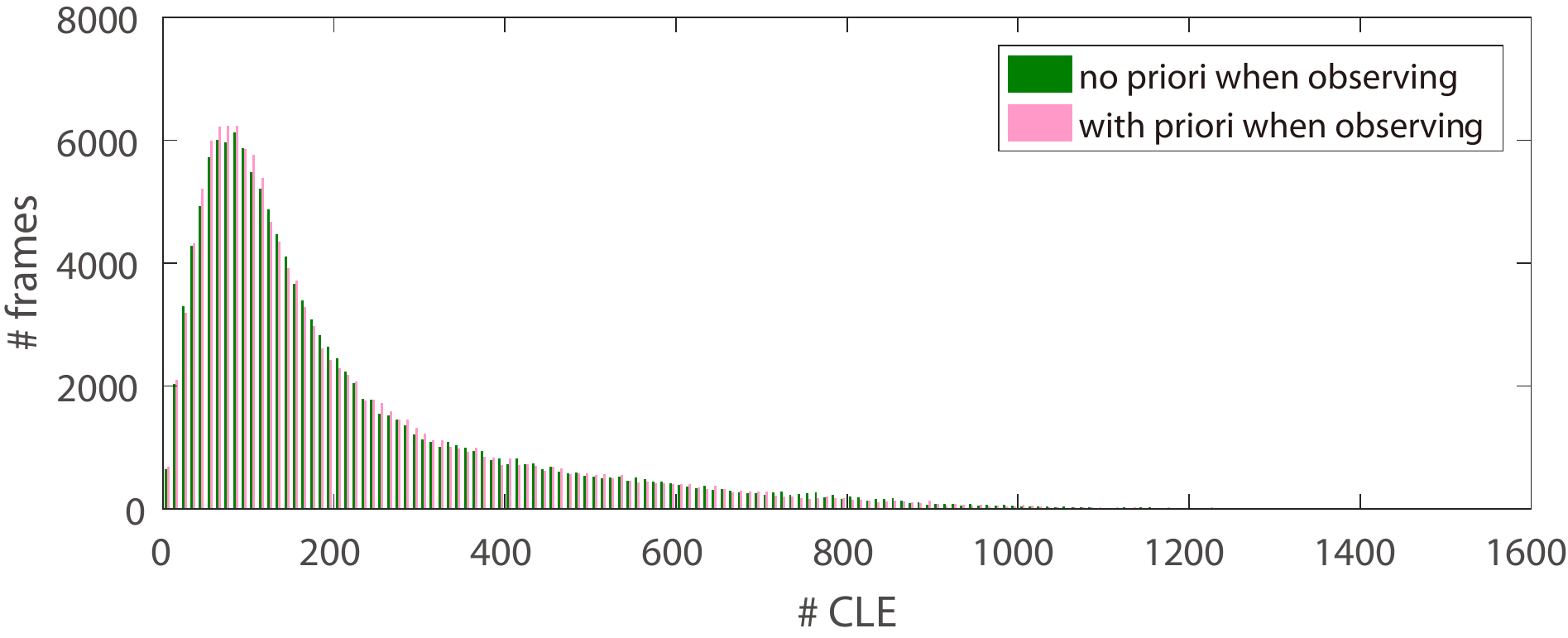}
    \caption{\small{The CLE distance distribution. y-axis denotes the frames with certain CLE range.}}
\label{fig8}
\vspace{-0.6cm}
\end{figure}

\subsection{Localization Error Analysis}
Beside the attended object category analysis, we further analyze the center location error (CLE) between the crash-object center with the peak point with the largest saliency value in our \emph{Type-1} attention maps. CLE represents the Euclidean distance between two points. Similarly, no-priori and with-priori are analysed. Differently, this analysis focuses on the abnormal situations (i.e, frames in AW).

\subsubsection{Overall Analysis}
\begin{figure}[t]
 \setlength{\abovecaptionskip}{0.0cm}
\centering
\includegraphics[width=0.7\linewidth]{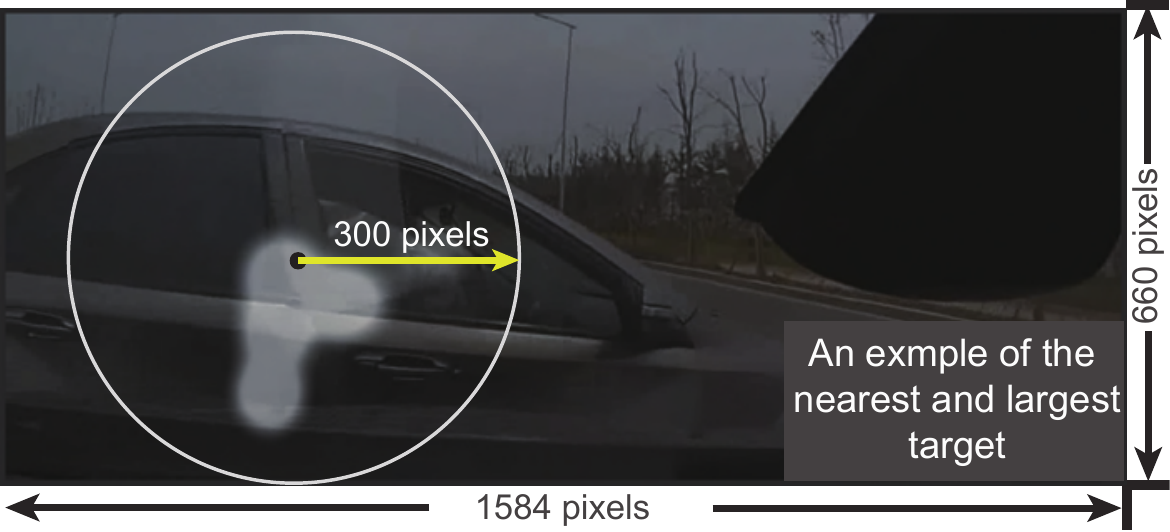}
    \caption{\small{An example with the largest crash-object. We can see that the crash car almost takes over a half of the image plane.}}
\label{fig9}
\end{figure}

We plot the histograms of CLE on all the frames in AW of our DADA-2000 dataset in Fig. \ref{fig8}, which again indicates that no-priori and with-priori have little influence on the crash-object localization. In addition, about $80\%$ percent of abnormal frames ($131,679$ frames in total) with smaller CLEs than $300$ pixels. Fig. \ref{fig9} gives an example for demonstrating the physical meaning of $300$ pixels. Actually, the size of the crash car is much larger than the circle with a radius of $300$ pixels. This analysis indicates that driver attention can play a positive role for crash-object localization to a large extent. There are two important evidences that: 1) the accident category of \emph{ego-car hitting car} takes the largest proportion in DADA-2000. When this kind of accidents occur, the radius of crash-object is prone to be larger than $300$ pixels, and 2) from Fig. \ref{fig3}, ego-car involved accident categories almost take half size of DADA-2000. Therefore, the CLEs smaller than $300$ pixels are acceptable. In addition, we also give an in-depth analysis for seeking the answer by checking the CLE evaluation in different sections in accident window.

\subsubsection{Analysis for Different Sections in AW}
Driver attention can localize the crash objects in accident to a large extent. Can it capture the crash-object when it just appears in the view? Following the temporal partition strategy of AW in Fig. ~\ref{fig5}, we plot the precision curves of different AW sections on all video clips in our DADA-2000. The precision curves contain two indicators: frame ratio and success rate, defined as follows:
\begin{itemize}
  \item Frame ratio is obtained by counting the number of frames in each AW section with lower CLE value than certain threshold compared to total frames in the same AW section. In this work, we analyze the frame ratio for different sections of AW in each video.
  \item Success rate denotes the video clip ratio which is obtained by counting the number of clips that over a half of ($\frac{1}{2}$) whose frames generate a lower CLE than the certain threshold, corresponding to the total clips (i.e., 2000). Similarly, success rate is analyzed for different sections of AW in each video.
\end{itemize}
\begin{table}[htpb]
\centering
\caption{The success rate ($\%$) when fixing the CLE lower than the distance threshold (DT) (pixels) for different sections in two sets of attention experiments, i.e., the start section with or without priori (star.wp and star.np), middle section with or without priori (mid.wp and mid.np), and end section with or without priori (end.wp and end.np).}
\begin{tabular}{|c||c|c|c||c|c|c|}

  \hline
  DT& Star.wp & Mid.wp & End.wp & Star.np & Mid.np & End.np \\
  \hline
  60 & 10.8 & 10.7 & 5.1 & 9.2&10.1&6.8 \\
 100 & 34.4 & 30.8 & 22.6 & 34.7&32.5&26.2 \\
160 & 59.0 & 57.2 & 49.1 & 61.1&59.0&52.1 \\
200 & 68.0 & 67.2 & 60.1 & \textbf{71.6}&69.6&64.0 \\
260 & \textbf{76.6} & \textbf{76.6} & \textbf{71.5} & 80.0&\textbf{79.5}&\textbf{73.3 }\\
300 & 80.0 & 81.3 & 76.6 & 83.5&83.8&79.1 \\
  \hline
\end{tabular}
\label{tab4}
\end{table}

\begin{figure}[t]
\centering
\includegraphics[width=\linewidth]{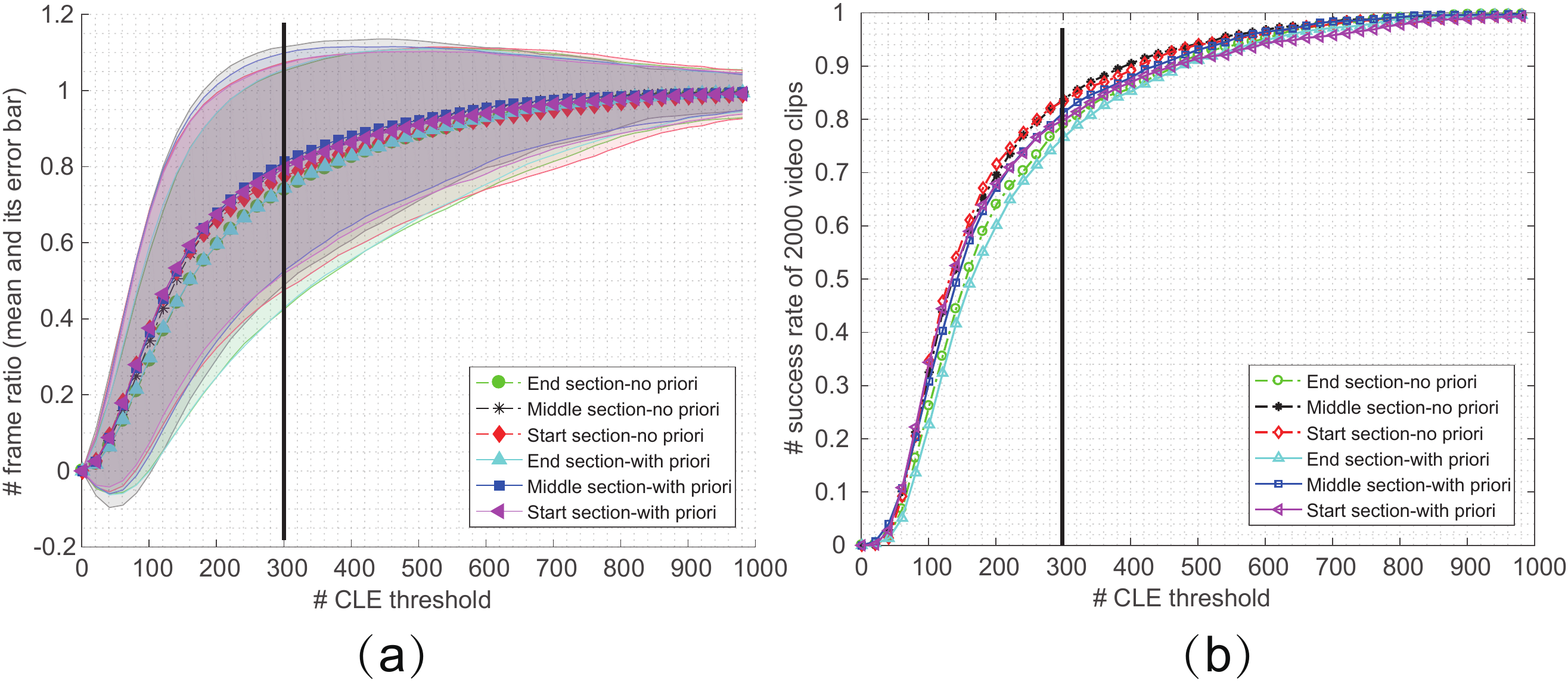}
    \caption{\small{The precision curves of driver attention localization for driving accident. x-axis specifies the CLE threshold. y-axis in (a) denotes the mean and its error bar (marked by the shadow area) of frame ratios of different AW sections. Instead, y-axis in (b) represents the success rate of all clips for different sections.}}
\label{fig10}
\end{figure}

 \begin{figure}[htpb]
 \setlength{\abovecaptionskip}{0.0cm}
\centering
\includegraphics[width=\linewidth]{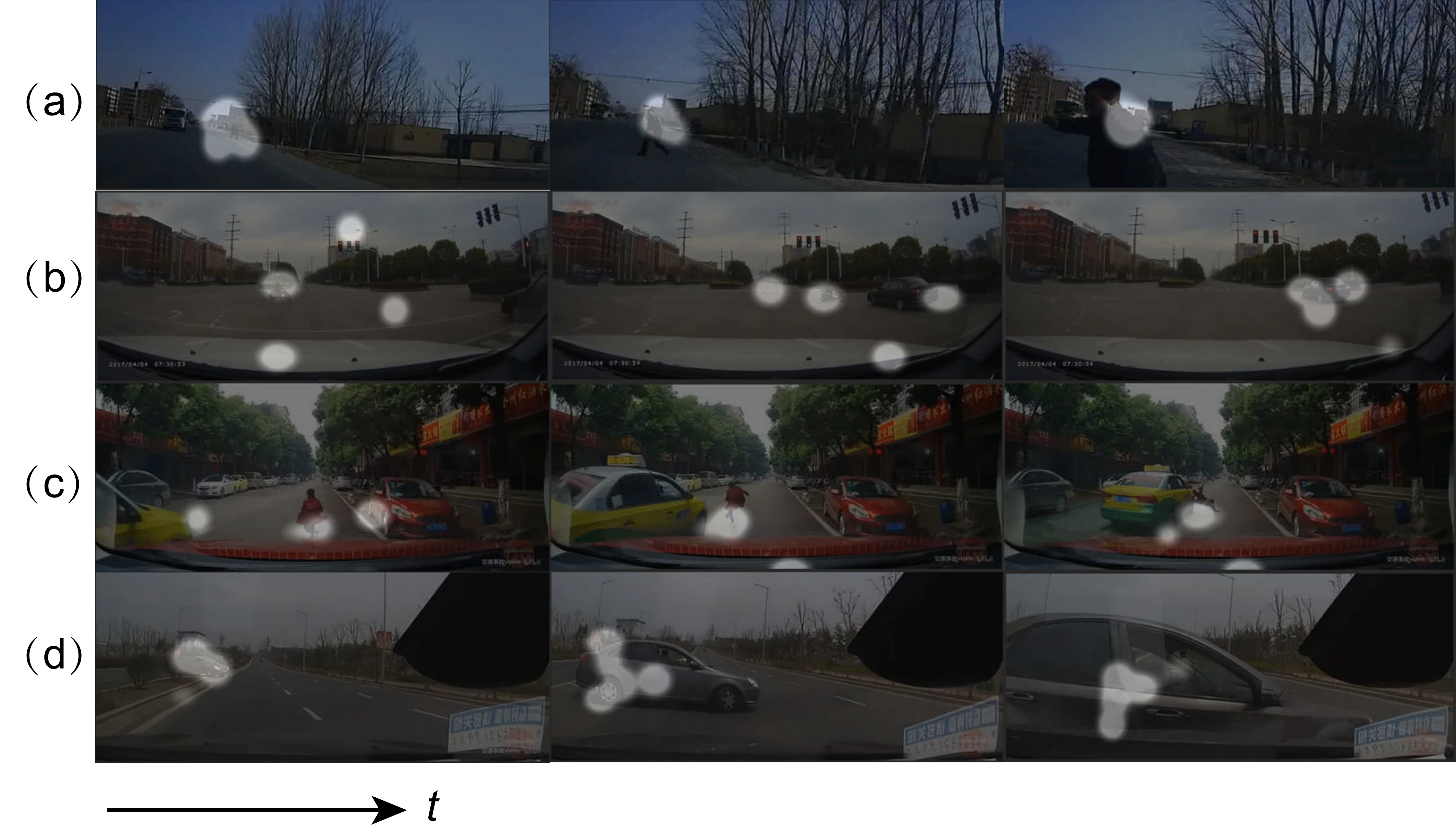}
    \caption{\small{\emph{Type1}-attention maps (five observers) of some typical accidents in DADA-2020. (a) a person crosses ahead of ego-car, (b) a car hits a motorbike, (c) a car hits a person, and (c) a car crosses ahead of ego-car.}}
\label{fig11}
\vspace{-0.6cm}
\end{figure}
The results are shown in Fig. \ref{fig10}(a) and Fig. \ref{fig10}(b), respectively. From these figures, we can see that with or without priori have a tiny influence for different AW sections. For different sections in AW, the mean frame ratio can reach $80\%$ when fixing the CLE threshold as $300$ pixels. This result is also found by the success rate. For a clearer demonstration, we present the success rate of different AW sections when fixing the CLE threshold as $60$, $100$, $160$, $200$, $260$, and $300$ pixels in Table. \ref{tab4}. From Table. \ref{tab4}, we can see that when fixing the CLE threshold as $260$ pixels, over $70\%$ video clips can be successfully attended. From these results, we can observe that driver attention is positive for crash-object localization in the accident window, especially for the start section of AW. In other words, driver attention can capture the object that will occur accident very early. Fig. \ref{fig11} demonstrates some typical accidents with the \emph{Type1}-attention maps. We can see that for the person category, driver attention usually focuses the head or foots, but not the center of the body. However, almost observers can capture the pedestrians or cars when they will occur accident.

\textbf{Highlights}: From these analysis, we can conclude that driver attention can obtain a \emph{positive help} for driving accident prediction. This will boost the future prediction model designing apparently. However, in complex scenes, the driver attention should be combined with more clues, such as the semantics and driving rules, to make a convincing prediction.

\section{Conclusions}

This work searched the answer for the question: can driving accident be predicted by driver attention? In order to answer this question, we construct a new and larger benchmark with $2000$ video clips (containing over $650,000$ frames) than the state-of-the-art related datasets. We named it as DADA-2000. For the construction of DADA-2000, we laboriously annotated the temporal accident window, spatial crash-object locations of all videos, and collect the driver attention data for each frame carefully. By quantitative analysis, we obtained a positive answer for this question.

\addtolength{\textheight}{-12cm}
{\scriptsize{
\bibliographystyle{IEEEtran}
\bibliography{bibfile}}
}

\end{document}